\documentclass[final,1p,times]{elsarticle}

\usepackage{amssymb}
\usepackage[utf8]{inputenc} 
\usepackage[T1]{fontenc}    
\usepackage{hyperref}       
\usepackage{url}            
\usepackage{booktabs}       
\usepackage{amsfonts}       
\usepackage{nicefrac}       
\usepackage{microtype}      
\usepackage{lipsum}		
\usepackage{graphicx}
\usepackage{natbib}
\usepackage{doi}

\usepackage{subfigure}
\usepackage{setspace}
\usepackage{amsmath, bm}
\usepackage{multirow}
\usepackage{multicol}
\usepackage[ruled, linesnumbered]{algorithm2e}

\usepackage{color}
\usepackage{lscape}

\newcommand{\blue}[1]{\textcolor{black}{#1}}
\newcommand{\red}[1]{\textcolor{black}{#1}}

\journal{Knowledge-Based Systems}

\begin{document}

\begin{frontmatter}

\title{Semi-supervised Training for Knowledge Base Graph Self-attention Networks on Link Prediction}

\author[1]{Shuanglong Yao}
\ead{shuanglongyao@gmail.com}

\author[1]{Dechang Pi\corref{cor1}}
\ead{pinuaa@nuaa.edu.cn}

\author[1]{Junfu Chen}
\ead{cjf@nuaa.edu.cn}

\author{Yufei Liu$^2$}
\ead{liuyufei@nuaa.edu.cn}

\author[1]{Zhiyuan Wu}
\ead{wuzhiyuan@nuaa.edu.cn}

\cortext[cor1]{Corresponding author}

\affiliation{organization={College of Computer Science and Technology, \\Nanjing University of Aeronautics and Astronautics},
            city={Nanjing},
            postcode={211106},
            country={China}}
\affiliation{organization={College of Information Engineering, \\Nanjing University of Finance and Economics}, city={Nanjing}, postcode={211106}, country={China}}

\begin{abstract}
The task of link prediction aims to solve the problem of incomplete knowledge caused by the difficulty of collecting facts from the real world.
\blue{
GCNs-based models are widely applied to solve link prediction problems due to their sophistication, but GCNs-based models are suffering from two problems in the structure and training process.
1) The transformation methods of GCN layers become increasingly complex in GCN-based knowledge representation models; 2) Due to the incompleteness of the knowledge graph collection process, there are many uncollected true facts in the labeled negative samples.
Therefore, this paper investigates the characteristic of the information aggregation coefficient (self-attention) of adjacent nodes and redesigns the self-attention mechanism of the GAT structure.
Meanwhile, inspired by human thinking habits, we designed a semi-supervised self-training method over pre-trained models. Experimental results on the benchmark datasets FB15k-237 and WN18RR show that our proposed self-attention mechanism and semi-supervised self-training method can effectively improve the performance of the link prediction task. If you look at FB15k-237, for example, the proposed method improves Hits@1 by about 30\%.}

\end{abstract}
\begin{keyword}
Knowledge graphs \sep link prediction \sep self-attention \sep semi-supervised self-training
\end{keyword}
\end{frontmatter}

\section{Introduction}
With the emergence and \red{more and more} \blue{applications} of knowledge graphs, knowledge graphs have been widely recognized by the industry \blue{as effectively improving} the performance of downstream tasks in the field of artificial intelligence, such as intelligence search, personalized recommendation, knowledge Q \& A\cite{Techniques, Applications}, etc.
However, collecting a complete knowledge graph from the real world is a \blue{huge task},
which requires \blue{a huge amount of} manpower\cite{Problems}.
Therefore, \blue{it's important to quickly and effectively fill in the gaps in the collected knowledge graph.}

\begin{figure}
  \centering
  \includegraphics[width=0.8\textwidth]{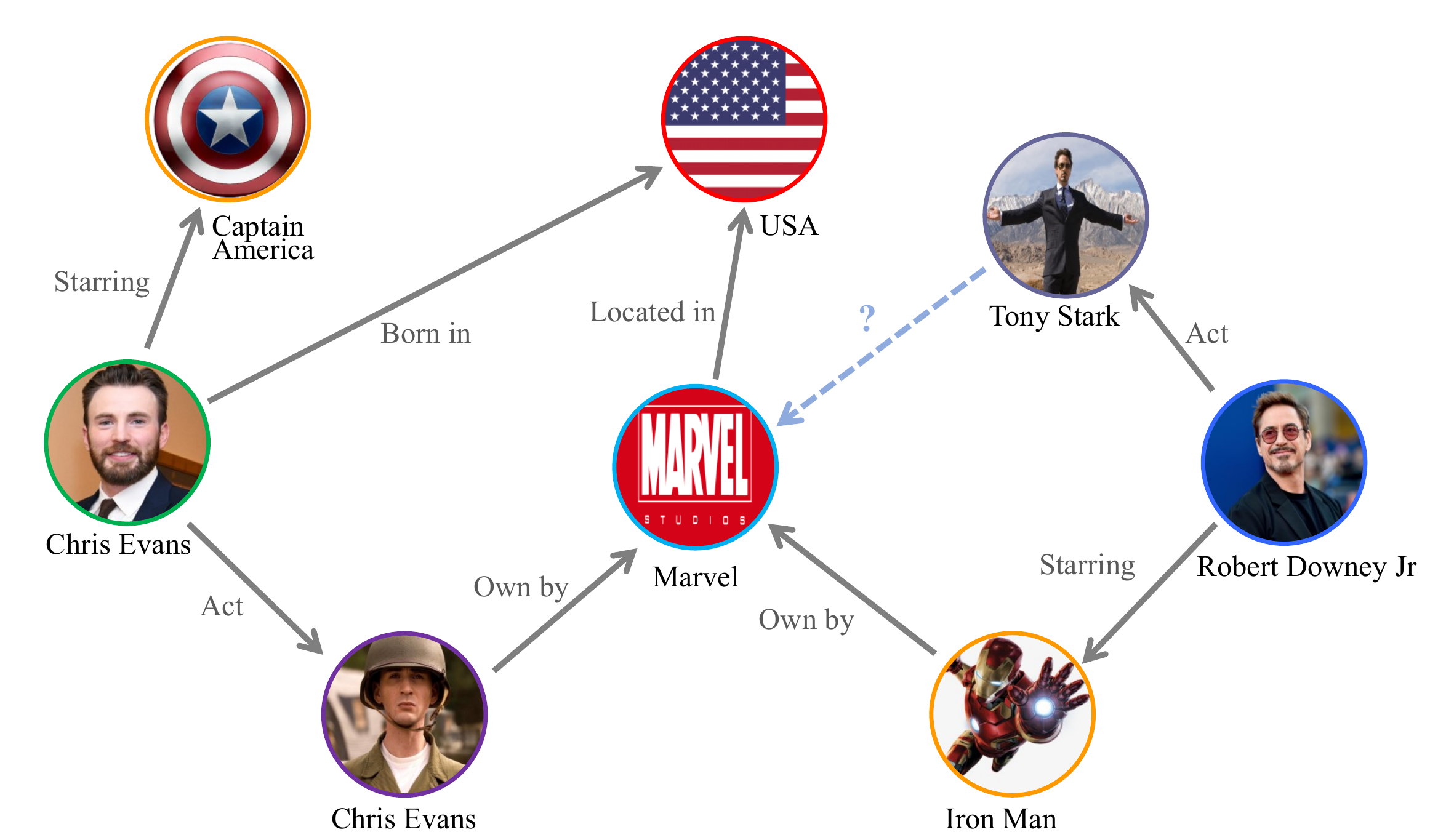}
  \caption{An example of knowlege graphs}
  \label{img1:KGs}
\end{figure}

Typically, a piece of knowledge fact is collected by knowledge graphs in the form of \blue{a} triplet, such as (head entity, relationship, tail entity), which \blue{is} mathematically expressed as $(h,r,t)$\cite{KGReview}. For each fact, the relationship connects the head entity and \blue{the} tail entity, and it is directional.
\blue{Knowledge completion is commonly referred to in knowledge graph research as ``link prediction'', which explains the missing triples through a mathematical model trained by the collected facts.}
In research of knowledge graph, knowledge completion is usually referred as link prediction\cite{Link_Prediction,Trans_Rules}, which reasons the missing triplet through mathematical model, which \red{can be} trained by the collected facts. Specifically, knowledge completion aims to infer the tail entity $t$ when the head entity $h$ and relationship $r$ are known, i.e., $(h,r,?)\to t$, or reason the head entity $h$ under the condition of given $r$and $t$, i.e., $(?,r,t)\to h$. This is very helpful \blue{for solving} reasoning problems in the real world. For instance, based on the knowledge graph shown in Figure \ref{img1:KGs}, the real question, “\emph{which company does the character Tony Stark belong to ?}”, can be simplified as a link prediction in the form of (Tony Stark, Own by, ?). As shown in Figure 1, the knowledge graph includes three facts: (Robert Downey Jr, act, Tony Stark); (Robert Downey Jr, Starring, Iron Man); and (Iron Man, Own by, Marvel), which implies the similar connection structure of neighbours of Chris Evans. Therefore, a new fact (Toney Stark, Own by, Marvel) can be roughly inferred. Obviously, applying the knowledge graph to \blue{knowledge} Q\&A will effectively improve the accuracy of answers.

Recently, \blue{GNN-based graph representation learning methods} have achieved remarkable results in many network tasks, such as node classification, node prediction, edge classification, \blue{link prediction,} and subgraph classification. As shown in Figure \ref{img1:KGs}, despite its heterogeneous directivity, the knowledge graph is also a kind of graph data.
Hence, it provides a new way to resolve the issue of knowledge incompletion through introducing the GNN-based graph representation approaches into the study of knowledge \blue{graphs}.
However, with the application of \blue{the} GNN-based model on the knowledge completion task, the structure of GNN-based knowledge representation models exhibits \blue{two obvious issues}: 1) For the GCN-based method, the complexity of linear transformation in GCN isn’t the key factor \blue{in} the overall improvement, which has been verified by experimental results of CompGCN; 2) For the GAT-based method, the attention vector captures \blue{much} global information noise in the process of \blue{neighbor} information aggregation.

Meanwhile, the negative sample generating strategy during training is also a key factor affecting the performance of the final model. In practice, the existing knowledge graphs only collect the true facts in the real world; that is, there is no negative samples that need to be generated by two kinds of strategies. 1) \textbf{Random generation}, \blue{replaces an entity from each triplet at random with a known entity from the knowledge graph's entity set}\cite{TransE}. In this method, the new triplets are generated with strong randomness and mislabeling. For example, the generated negative samples may be the collected facts in the test set. 2) \textbf{Direct generation}, directly defines each uncollected fact as a negative sample, which \blue{leads to} higher mislabeling, but this way is simple and \blue{generates} quickly\cite{ConvE}. Obviously, no matter which method is used, there will inevitably be a large number of “negative samples” with \blue{the} wrong labels.

\blue{This paper proposes a new \textbf{\emph{knowledge base graph self-attention network}} (KBGSAT) by redesigning the self-attention mechanism of GATs in order to address the structural problem and achieve more efficient performance. Firstly, the simple splicing operation is used as the linear transformation in this paper. Secondly, unlike traditional GAT-based models, which focus on the direct setting of the global attention vector, this paper studies the softmax influence of link features obtained by the linear transformation and the multiplication operation. Theoretically, they are the key factors affecting softmax in the attention mechanism.}

Inspired by the associative memory process of human brain, this paper proposes a novel \textbf{\emph{semi-supervised self-training method}} for \blue{the fine-tuning process}. Associative memory is a method of human learning through associative comparison in the process of memory. It can be regarded as a retraining process of the model through new \blue{facts generated} by the pred-trained model. Still, taking the knowledge graph of Figure \ref{img1:KGs} as an example, it is difficult \blue{to} \red{directly} \blue{reason} the question (Robert Downey Jr, Born in, ?). \blue{When comparing and associating Robert Downey Jr with Charles Evans, it is easy to infer (Robert Downey Jr, born in USA) if two predicted facts (Tony Stark, owned by Marvel) and (Captain America, owned by Marvel) are added to the current knowledge graph of Figure} \ref{img1:KGs}. Actually, the facts, which are generated during the process of associative memory, are not necessarily right, but they can still improve people’s reasoning \blue{abilities}. Furthermore, this process is extraordinary for knowledge completion, because models can be retrained on \blue{meaningful questions} with unknown answers through this method. In other words, the question itself can \blue{be used} to train the model.

In order to evaluate the effectiveness of the proposed model KBGSAT and semi-supervising training method, we conducted experiments on the widely used benchmark data sets FB15k-237 and WN18RR. The experimental results show that the model KBGSAT, \blue{with semi-supervised self-training}, has achieved better performance than the state-of-the-art link prediction approaches on some metrics. \blue{In particular, it has achieved a relative improvement of about 30.7\% in Hits@1 on the benchmark data FB15k-237.}

Our specific contributions are summarized as follows:
\begin{itemize}
\item[1)] Propose a new knowledge completion model in \blue{graph} self-attention networks\blue{, which} replaces the translational attention vector by the line transformation of \blue{a soure} entity, and redesign the self-attention mechanism in this model.
\item[2)] Inspired by the memory process of \blue{humans}, we \blue{designed} a novel semi-supervised self-training method for retraining process. This approach introduces some new unverified positive samples, which are generated by the pretrained model, to improve the \blue{performance} of the model.
\item[3)] Evaluate the proposed KBGSAT and the semi-supervised self-training method on two recent standard link prediction \blue{datasets,} FB15k-237 and WN18RR.  Experimental results show that the model and approach can effectively improve the performance on the link prediction tasks.
\end{itemize}

\blue{This paper is mainly divided into five sections. Section 2 introduces related works on the link prediction (knowledge completion) task; Section 3 details the structure of the model KBGSAT and the semi-supervised self-training algorithm; Section 4 depicts the experimental process and results; and Section 5 summarizes the paper and suggests future work.}

\section{Related Work}

With the development of knowledge graphs, knowledge completion has attracted more and more attention. Given a knowledge \blue{graph,} $\mathcal{G}=(\mathcal{E},\mathcal{R},\mathcal{T})$, where $\mathcal{E}$ defines \blue{the node set} containing all entities, $\mathcal{R}$ defines the label set of relationships, and $\mathcal{T}$ represents the facts in the form of \blue{a triple}.
\blue{Each piece of real-world information is represented as a triple $(h, r, t)$ in $\mathcal{T}$, where $h$, $t$ represent the head and tail entities, respectively, and $r$ describes the directed relationship between the two entities.}
As mentioned in the premature, the link prediction models complete the missing triple based on the collected facts. According to \cite{Link_Prediction}, the current mainstream link prediction models are divided into three categories:

\textbf{Tensor Decomposition Models}

Among such as a tensor decomposition model, a knowledge graph is considered as \blue{a} 3D adjacency matrix, and the link prediction problem is described as a tensor decomposition task, in which tensors are decomposed into low dimensional vectors, which are defined as the embeddings of entities and relationships. According to whether the bilinear product operation is used, tensor decomposition models are divided into 2 \blue{classes}: (1) Bilinear models, which generate various \blue{extended} models by adding specific constraints to the embeddings, such as DistMult\cite{DistMult}, ComplEx\cite{ComplEx}, Analogy\cite{Analogy} and SimplE\cite{SimplE}; (2) Non-bilinear models, which combine the embeddings of entities and relationships and build models by different non-bilinear operations, such as HolE and TuckER\cite{TuckER}.

\textbf{Geometric Models}

The geometric models interpret the relationship between two entities in the knowledge graph as a potential spatial transformation. Hence, the score of each fact is calculated through norm 1 or 2 on the distance between the tail embedding vector and the transformation feature, which \red{can be} obtained by the head entity and the relationship after the spatial transformation. Meanwhile, geometric models are also known \blue{as translation} models\blue{. For} example, TransE\cite{TransE} regards each relationship \blue{as a} “translation” from head entities to tail entities, and “translation” is represented by the addition operation in space, ${{\bm{v}}_{h}}+{{\bm{v}}_{r}}\to {{\bm{v}}_{t}}$. Derived from TransE via introducing additional information features, StransE\cite{StransE} and CrossE\cite{CrossE} \blue{achieve better} accuracy. Different from the above models, TorusE\cite{TorusE} and RotatE\cite{RotatE} add a rotation operation during spatial transform, TransRHS\cite{TransRHS} is built with the relation hierarchical structure.

\textbf{Deep Learning Models}

\blue{Compared} to the two kinds of previous models, the deep learning models \blue{are typically built} by one or multiple layers of neural networks, which consist of some complex operations, such as convolution layers, recurrent layers, or graph convolutional layers. Based on the neural architecture, we identify three \blue{groups} in this family: (\emph{i}) employing convolutional neural networks, such as ConvE\cite{ConvE}, ConvR\cite{ConvR}, and HMAE\cite{HMAE}; (\emph{ii}) \blue{employing} recurrent neural networks, RSN\cite{RSN}; (\emph{iii}) \blue{employing} graph convolutional networks, such as R-GCN\cite{R-GCN}, CompGCN\cite{CompGCN}.

\section{KBGSAT and semi-supervised self-training method}

In this section, we \blue{first analyze} the structure of \blue{GATs and then} introduce the knowledge base graph self-attention networks (KBGSAT) for the link prediction of knowledge graphs. Finally, we present the semi-supervised self-training method for fine-tuning this \blue{model,} KBGSAT.

\subsection{GATs}

\begin{figure}
  \centering
  \includegraphics[width=0.95\textwidth]{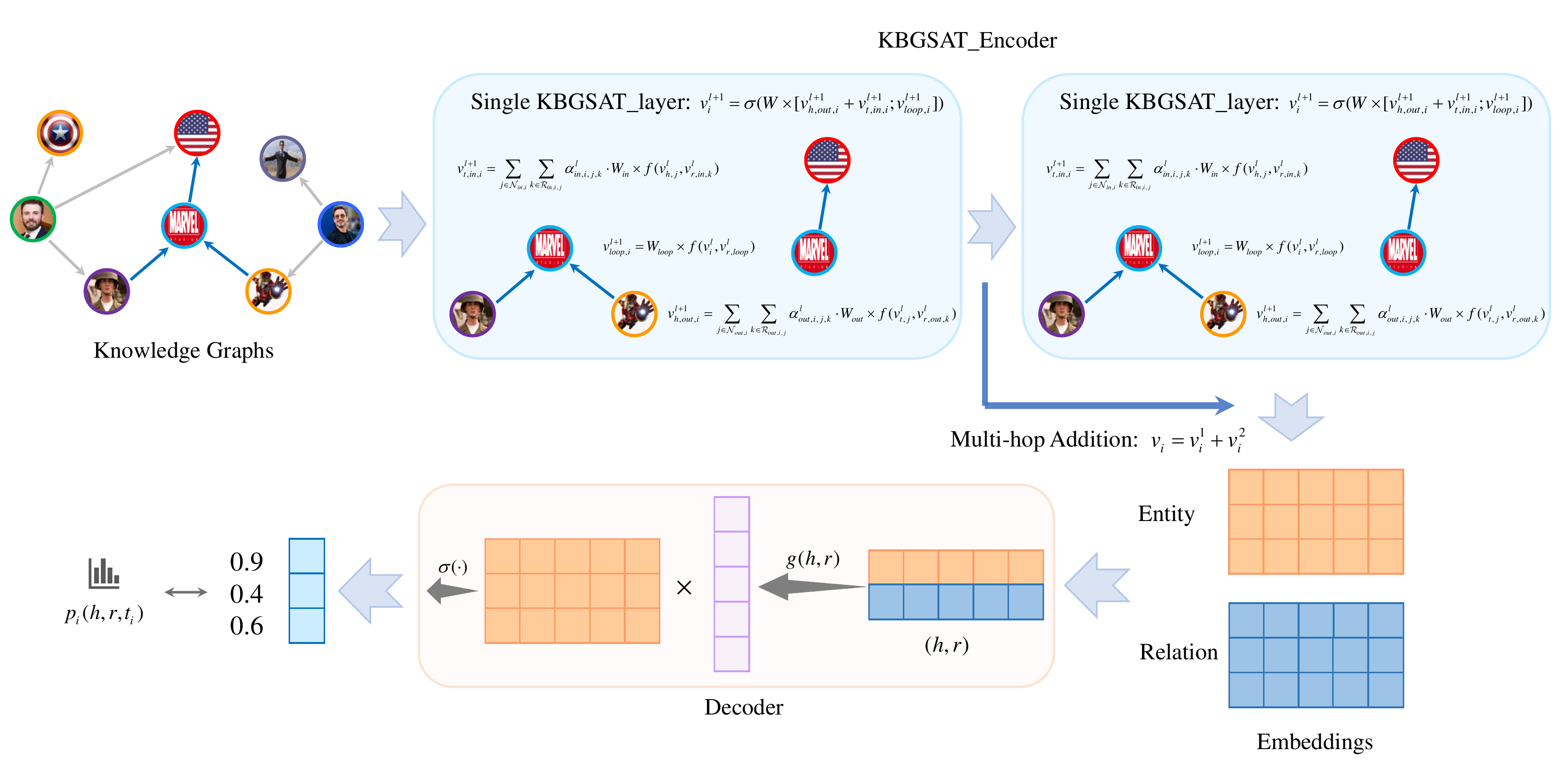}
  \caption{The architecture illustration of KBGSAT with two graph convolutional layers.}
  \label{img2:model}

\end{figure}

The purpose of GATs is to solve the graph node representation task on isomorphic network. The core formula is summarized as the following formulas\cite{Link_Prediction}:
\begin{equation}
  \bm{v}^{l,*}\leftarrow \sigma (\bm{W}^l\times \bm{v}^l)
  \label{eq:1}
\end{equation}
\begin{equation}
  \bm{v}_{i}^{l+1}\leftarrow \text{Aggregate}(\{\bm{v}_{j,}^{l,*},\forall j\in \{i\}\bigcup {{\mathcal{N}}_{i}}\})
  \label{eq:2}
\end{equation}
where $\times$ is the matrix multiplication operation, $l$ represents the index of the convolution layer in the model, then $\bm{v}_{i}^{l}$ describes the node $i$ input feature vector of $l$-th layer, and $\bm{v}_{i}^{l+1}$ defines the node $i$ output feature vector of $l$-th layer. \blue{Furthermore}, $\mathcal{N}_i$ defines the neighbor set of the network node $i$, $\text{Aggregate}(\cdot)$ is a kind of aggregation functions, $\bm{W}$ represents the transformation matrix, and $\sigma(\cdot)$ is the activation function. Formula \ref{eq:2} captures the aggregation of information from \blue{the neighbors of a node}, while formula \ref{eq:1} applies the linear transformation and activation function to integrate the aggregated information into a better embedding vector. Actually, different from isomorphic networks, knowledge graphs are the more complex and variable \blue{kinds} of heterogeneous networks. In a knowledge graph, the nodes are divided into different categories, and the edges are also separated into a variety of categories. Therefore, solving the complex problems caused by heterogeneous relationships is the key technology to overcome when \blue{GAT-based} models are applied to knowledge representation. Unlike the proposed \blue{GAT-based} models, this paper designs a new self-attention mechanism for the GATs framework. As shown in Figure \ref{img2:model}, the architecture of KBGSAT is mainly divided into two parts: 1) GATs-Encoder\blue{ generates embedding} representations of entities and relationships in knowledge graphs; 2) KB-Decoder\blue{ predicts links based on the embeddings obtained}.

\subsection{GATs-Encoder}

\blue{The He-initializer approach randomly initializes two GATs-Encoder input feature matrices: the initial entity feature matrix $\bm{E}\in {{\mathbb{R}}^{|\mathcal{E}|\times d}}$ and the initial relationship feature matrix $\bm{R}\in {{\mathbb{R}}^{|\mathcal{R}|\times d}}$.}
$|\mathcal{E}|$, $|\mathcal{R}|$ represent the number of entities and relationships in the knowledge graphs, respectively. Meanwhile, $d$ defines the dimension of embedded vectors. Inspired by CompGCN \cite{CompGCN}, we divide the single-layer entity aggregation process into three parts: the out-relationship process, the in-relationship process, and the self-loop process. Hence, the output embedding of entity $i$ on the $l$-th layer is given by:
\begin{equation}
  \bm{v}_{h,out,i}^{l+1}=\sum\limits_{j\in {{\mathcal{N}}_{out,i}}}{\sum\limits_{k\in {{\mathcal{R}}_{out,i,j}}}{\alpha _{out,i,j,k}^{l}\cdot {{W}_{out}}\times f({\bm{v}_{t,j}^{l}},\bm{v}_{r,out,k}^{l})}}
\label{eq:3}
\end{equation}
\begin{equation} \bm{v}_{t,in,i}^{l+1}=\sum\limits_{j\in {{\mathcal{N}}_{in,i}}}{\sum\limits_{k\in {{\mathcal{R}}_{in,i,j}}}{\alpha _{in,i,j,k}^{l}\cdot {{W}_{in}}\times f(\bm{v}_{h,j}^{l},\bm{v}_{r,in,k}^{l})}}
\label{eq:4}
\end{equation}
\begin{equation} \bm{v}_{loop,i}^{l+1}={{W}_{loop}}\times f(\bm{v}_{i}^{l},\bm{v}_{r,loop}^{l})
\label{eq:5}
\end{equation}
\begin{equation}
  \bm{v}_{i}^{l+1}=\sigma (\bm{W} \times [\bm{v}_{h,out,i}^{l+1};\bm{v}_{t,in,i}^{l+1};\bm{v}_{loop,i}^{l+1}])
\label{eq:6}
\end{equation}

\begin{equation}
 \bm{v}_r^{l+1}=\bm{W}_r\times\bm{v}_r^l
\label{eq:7}
\end{equation}
where $\cdot$ is the scalar multiplication\blue{ and the subscripts} $out$, $in$ \blue{and} $loop$ \blue{denote the out-relationship,} in-relationship and self-loop, respectively. \blue{For example,} in formula \ref{eq:3}, $\bm{v}_{h,out,i}^{l+1}$ represents the acquired embedding vector of entity $i$ based on out-relationship on the $l$-th layer, $\bm{W}_{out}$ \blue{represents} the linear transformation matrix of out-relationship aggregations, $\alpha _{out,i,j,k}^{l}$ is the out-relationship aggregation coefficient, $\mathcal{N}_{out,i}$ defines the set of \blue{neighbor nodes} based on the out-relationship of the entity $i$, $\mathcal{R}_{out,i,j}$ describes the out-relationship set between the head entity $h_i$ and the tail entity $t_j$, $r_{out,k}\in\mathcal{R}_{out,i,j}$.
let $f(x,y)=[x;y]$, $[\cdot;\cdot]$ define the splicing operation. For consistency, let $\bm{W}_{out},\ \bm{W}_{in},\ \bm{W}_{loop}\ \in \ \mathbb{R}^{d\times 2d}$ and $\bm{W}\in  \mathbb{R}^{d\times 3d}$.

In traditional \blue{GAT-based} models, such as KBGAT, the aggregation coefficient is calculated by:
\begin{equation} \alpha _{i,j,k}^{l}=\frac{\exp ({\bm{a}^{\top }}[\bm{v}_{h,i}^{l};{\bm{v}_{r,k}^l};\bm{v}_{t,j}^{l}])}{\sum\nolimits_{j'\in {{\mathcal{N}}_{i}}}{\sum\nolimits_{k'\in {{\mathcal{R}}_{i,{j}'}}}{\text{exp}({\bm{a}^{\top }}[\bm{v}_{h,i}^{l};{\bm{v}_{r,{{k}'}}^l};\bm{v}_{t,{{j}'}}^{l}])}}},
\label{eq:8}
\end{equation}
where $\bm{a}$ represents the independent self-attention vector. The attention vector $\bm{a}$ is a global vector and it captures the information propagation characteristics of each fact during the training process.

\blue{In contrast to} the traditional \blue{GAT-based} models, we redesign the self-attention mechanism. In this paper\blue{, the} splicing operation is no longer used in the self-attention mechanism, and the independent global attention vector is removed. For the out-relationship, the specific calculation process of the aggregation coefficient $\alpha _{out,i,j,k}$ is computed by the following formula:
\begin{equation}
  \alpha _{out,i,j,k}^{l}=\frac{\text{exp}({{({\bm{W}}_{out}' \times\bm{v}_{h,i}^{l})}^{\top }}(\bm{W}_{out}\times f(\bm{v}_{t,j}^{l},{\bm{v}_{r,out,k}})))} {\sum\nolimits_{j'\in {{\mathcal{N}}_{out,i}}}{\sum\nolimits_{k'\in {{\mathcal{R}}_{out,i,{j}'}}}{\text{exp}({{({\bm{W}}_{out}' \times\bm{v}_{h,i}^{l})}^{\top }}(\bm{W}_{out}\times f(\bm{v}_{t,j'}^{l},{\bm{v}_{r,out,k'}})))}}}.
\label{eq:9}
\end{equation}
\blue{In comparison to the formula} \ref{eq:8}, the independent global attention vector is replaced by the transformation feature ${\bm{W}}_{out}' \times\bm{v}_{h,i}^{l}$ of the head entity, $\bm{W}_{out}'\in \mathbb{R}^{d\times d}$. So that, \blue{when} calculating the aggregation coefficient, the molecular differences in softmax operation of each head entity only come from relationships and tail entities.
Although the above changes have reduced the ability of $\alpha$ to capture global information, the extracted information from \blue{neighbors has} become purer during training, \blue{and the} whole calculation process has become more consistent with the structure of the Transformer.

Further, in order to overcome the loss of information caused by multi-layer convolutional networks, this paper used the Skipped mechanism, in which the final embedding \blue{representations} of entities are obtained by adding the output features from two-layer graph convolutional networks, $\bm{v}_i=\bm{v}_i^1+\bm{v}_i^2$.

\subsection{KB-Decoder}
In order to verify the effectiveness of \blue{GATs-encoder} structure, this paper applies TransE, DistMult and ConvE as the score \blue{functions} $\phi(\cdot)$ of \blue{the decoder}.
Meanwhile, given the condition pair $(h,r)$, the sigmoid function $\sigma(\cdot)$ is employed to calculate the probability $p_i=\sigma(\phi(h,r,t_i))$ of there being a link $r$ to the tail entity $t_i$.

When \blue{the decoder} is TransE, $\phi(\bm{v}_h,\bm{v}_r,\bm{v}_{t,i})=;\bm{v}_h+\bm{v}_r-\bm{v}_{t,i};_{\mathcal{l}_1}$;

When  \blue{the decoder} is DistMult, $\phi(\bm{v}_h,\bm{v}_r,\bm{v}_{t,i})=\bm{v}_h\times \bm{v}_r \times \bm{v}_{t,i}$, and $\bm{v}_r$ is diagonal;

When \blue{the decoder} is ConvE, $\phi(\bm{v}_h,\bm{v}_r,\bm{v}_{t,i})=g(\bm{W}\times g([\bm{v}_h;\bm{v}_r]\otimes \omega )+b)\times \bm{v}_{t,i}$.

Where, $\times$ represents the multiplication of the matrix, $\otimes$ is the convolution operation, $\omega$ defines the convolution kernel and $g$ is the activation function, respectively. To train the parameters of the KBGSAT model, minimize the binary \blue{cross-entry} loss:
$\mathcal{L}(p,q)=-\frac{1}{N}\sum\limits_{i}{({{q}_{i}}\cdot \log (}{{p}_{i}})+(1-{{q}_{i}})\cdot \log (1-{{p}_{i}}))$.
Similar to ConvE, this paper use 1-N score label.

\subsection{Semi-supervised self-training method}
In real life, people will make preliminary associative inferences (the process of generating unverified facts \blue{according to} the trained model) for unknown problems based on \blue{their} experience memory (the trained model). \blue{Meanwhile,} these associations in turn continuously optimize the experience memory (retaining the model). Inspired by this \blue{associative} memory approach, a novel semi-supervised self-training method is designed in this paper.

Different from the traditional training process, the semi-supervised self-training method proposed in this paper is divided into two parts: \blue{1) the training phase; 2) the generation of new facts}. During training process, the directly generated sampling method is applied for the overall training of model parameters. In the new fact generation, the scores are sorted in descending sequence by the current trained model, and the node with the best score is taken as the prediction node. \blue{In the} sorting process, the scores of train facts have been removed from the predicted scores. Obviously, the new fact generation process is an application of the \blue{best-neighbor} algorithm \blue{to} the link prediction task.

\begin{algorithm}
  \caption{Semi-supervised self-training method}
  \label{alg:1}
  \KwIn{Original training set $\mathcal{T}$; pretrained model KBGSAT.}
  \KwOut{Final trained KBGSAT.}

  Initialized the $l$-th generated triples $\mathcal{T}'=\text{None}$.\\
  Generate the new triples $\mathcal{T}_{new}$ by Algorithm 2 based on the pretrained model.\\
  $\mathcal{T}_{new}=\mathcal{T}\cup \mathcal{T}_{new}$ \\
  \% The KBGSAT is trained on 300 epochs. In the code, we have employed the early stop strategy.\\
  For $i=1$ to 300 do:\\
       \qquad \% \emph{The batch pairs and labels are generated by the traditional direct generation sampling.} \\
       \qquad For batch\_pairs, batch\_labels in generate\_batch\_data($\mathcal{T}_{new}$) do:\\

       \qquad \qquad \% \emph{$W_e$, $W_r$ are the embeddings of entities and relations, respectively.}\\
       \qquad \qquad $W_e,\ W_r=\text{GATs\_Encoder}(\mathcal{T}_{new})$;\\
       \qquad \qquad For $(h,r)$, $q$ in batch\_pairs, batch\_labels do:\\
       \qquad \qquad \qquad $\bm{v}_h=\text{Index\_select}(W_e,\ h)$;\\
       \qquad \qquad \qquad $\bm{v}_r=\text{Index\_select}(W_r,\ r)$;\\
       \qquad \qquad \qquad $\bm{v}_{t,i}=\text{Index\_select}(W_e,\ t_i)$;\\
       \qquad \qquad \qquad $p_i=\sigma (\phi(\bm{v}_h,\ \bm{v}_r,\ \bm{v}_{t,i}))$;\\
       \qquad \qquad $\mathcal{L}(p,\ q)=-\frac{1}{N}\sum_i(q_i\cdot\log(p_i)+(1-q_i)\cdot\log(1-p_i))$;\\
       \qquad \qquad Unpdate the parameters of the model KBGSAT;\\
       \qquad \qquad End;\\
       \qquad End;\\
       \qquad If convergence do:\\
        \qquad \qquad Break;\\
      \qquad End;\\
       End;\\
\end{algorithm}

\begin{algorithm}
 \caption{Generate new triples}
 \label{alg:2}
 \KwIn{The pretrained KBGSAT; the original training set $\mathcal{T}$; the condition pairs $\mathcal{K}$ (which are generated by removing the head or tail of each triple in valid or test set, i.e., for $\forall(h,r,t)\in\mathcal{T}_{new}$, $\exists(h,r)\  \text{and}\ (t, r+|\mathcal{R}|)\in \mathcal{K}$).}
 \KwIn{The new triples for training.}
 Let $\mathcal{T}_{new}$ be initialized as an empty set.\\
 For $(h,r)\ \text{in}\ \in \ \mathcal{K}$ do:\\
 \qquad $p=\text{KBGSAT}(h,r)$;\\
 \qquad \% Extract corresponding index as predicted tails by descending.\\
 \qquad $\mathcal{T}_{pred}=\text{Argsort}(p)$;\\
 \qquad For $t_{pred,\ i}\ \in \ \mathcal{T}_{pred}$ do:\\
 \qquad \qquad If $(h,r,t_{pred,\ i})\ \notin \mathcal{T}$ do:\\
 \qquad \qquad \qquad Add the new triple $(h,r,t_{pred,\ i})$ into $\mathcal{T}_{new}$;\\
 \qquad \qquad \qquad Break;\\
 \qquad \qquad End;\\
 \qquad End;\\
 End;\\
\end{algorithm}

As shown in Algorithm \ref{alg:1}, the key technology of our semi-supervised self-training method is to generate unverified positive samples by adding a generation algorithm before training.
In Algorithm \ref{alg:2}, the core of generating new facts is a greedy strategy (manifesting on the network, extracting the best \blue{neighbors}). The core \blue{step is as follows}: in line 3, scores of the condition pair on the entity set are calculated by the current trained model; in line 4, take the corresponding index as the tail entity ID and then sort the scores in descending order.

It is worth noting that using the condition pairs \blue{from a} verification or test set is essentially different from directly applying the verification or test set to generate negative samples in the training process. We take the human reinforcement learning process as an example. People apply \blue{their} past verified experience to build a specific knowledge memory (the model parameters converge after training on the training set). When encountering a new reasoning problem (the new \blue{problem is} usually \blue{short of} two elements \blue{in a triple}, such as in the introduction (Captain America, Own by, ?)), people will make associative inferences (true or false without verification) based on the learned memory. Meanwhile, the unverified inferences often imperceptibly affect people’s constructed knowledge memory in real life. Hence, the associative memory process is a part of people’s thinking when \blue{facing new problems}. By analogy with the above reasoning process, the semi-supervised self-training is also regarded as \blue{an} “associative memory process” before making the final prediction.

\section{Experiments}
This section introduces the details of the experiment, which is mainly divided into five parts: 1) experimental benchmark data, 2) comparison models, \blue{3) evaluation} metrics, 4) experimental settings\blue{, and 5)}experimental results.

\subsection{Experimental data}
In order to fairly and comprehensively verify the effectiveness and scalability of the model KBGSAT, experiments were done on two widely used public data sets:

\textbf{FB15k-237} is a subset built by Toutanova and Chen to solve the test leakage problem on FB15k, in which a large number of facts have a reciprocal or equivalent relationship\blue{. For} example, if there is a triple $(h,r,t)$, the triple $(h,-r,t)$ is also collected, where$r$ and $-r$ are reciprocal relationships. This problem has been demonstrated by Toutanova and Chen to achieve advanced results by building a simple model based on \blue{these} reciprocal or equivalent relationships on FB15k \cite{ConvE}. \blue{To avoid this problem, FB15k-237 is made by only including facts which have filtered an equal or reciprocal relationship.}

\textbf{WN18RR} is a subset of WN18, which is \blue{a} widely studied subnetwork of the English lexical semantic network WordNet. Similar to FB15k, Dettmers et al. find that WN18RR also has the reverse relationship and test leakage problem. Therefore, Dettmers et al. filter out the facts with reverse or equivalent relationships in the WN18 to obtain the dataset WN18RR through the method of Toutanova et al \cite{Link_Prediction}.

FB15k-237 and WN18RR data statistics parameters are shown in Table \ref{tab:statistic}.

\begin{table}[h]
  \footnotesize
  \centering
	\caption{Statistical parameters of data sets}
	\begin{tabular}{lllllll}
		\toprule
		Data	&Entity	&Relation	&Train	&Valid	&Test \\
		\midrule
		FB15k-237	&14541	&237	&272115	&17535	&20466	 \\
		WN18RR	&40943	&11	&86835	&3034	&3134	 \\
		\bottomrule
	\end{tabular}
	\label{tab:statistic}
\end{table}

\subsection{Comparison models}
\blue{We compare three groups of knowledge representation models on the two datasets mentioned above.} Three \blue{group} models are listed as follows:

\textbf{1. Tensor Decomposition Models}

\textbf{DistMult}\cite{DistMult} reduces the model’s parameters and improves the training speed by embedding the relationships into the angular matrix.

\textbf{ComplEx}\cite{ComplEx-N3} is the extension model of DistMult on the complex space, but the double linear operation is replaced by the Hermitian operation.

\textbf{Analogy}\cite{Analogy}, of which the core idea is to employ the commutativity constraint of matrix to model analogy reasoning attribute like the parallelogram, adds two constraints to the bilinear score function: for each relationship, $\bm{v}_r\times\bm{v}_r^{\top}$, and $\bm{v}_r$ is a normal matrix; for each relationship pair, $r_1$, $r_2$, $\bm{v}_{r_1}\circ \bm{v}_{r_2}=\bm{v}_{r_2} \circ \bm{v}_{r_1}$.

\textbf{SimplE}\cite{SimplE} is a tensor decomposition method, based on the CP method, to solve the independence problem of the head entity and the tail entity of each triple during training. SimplE embeds any entity into two independent representation vectors, and each relationship into two independent diagonal matrices.

\textbf{HolE}\cite{HolE} obtains the scores of facts through calculating the matrix multiplication between the relationship embeddings with the cycle correlation of the embedding vectors of both the head entity and the tail entity.

\textbf{TuckER}\cite{TuckER} decomposes a third-order tensor into a core tensor based on \blue{the} tucker decomposition. \blue{In TuckER, the entity embedding vector is represented by $\bm{v}_e\in \mathbb{R}^{d_e}$, the relationship embedding vector is described by $\bm{v}_r\in\mathbb{R}^{d_r}$, and the embedding dimensions of entities and relationships are defined by $d_e$ and $d_r$, respectively.}

\textbf{2. Geometric Models}

\textbf{TransE}\cite{TransE} is inspired by Word2Vec, and \blue{considers} the relationship \blue{to be a} “translation” from the head entity to the tail entity, \blue{i.e.} ${{\bm{v}}_{h}}+{{\bm{v}}_{r}}\to {{\bm{v}}_{t}}$.

\textbf{CrossE}\cite{CrossE} is the most recent and effective translational model with additional relation-specific embeddings $\bm{c}_r$ and uses element-wise products to combine $\bm{v}_h$ and $\bm{v}_r$ with $\bm{c}_r$.

\textbf{RotatE}\cite{RotatE} represents the relationship of each fact as a rotation in the complex space and applies the rotation to the head entity via the element-wise operation.

\textbf{3. Deep learning Models}

\textbf{ConvE}\cite{ConvE} combines and reconstructs the embeddings of the head entity and relationship into the input matrix, then employs a set of low-dimensional convolution kernels to obtain a feature.

\begin{landscape}
\begin{table}
  \renewcommand\arraystretch{1.5}
  \footnotesize
  \centering
	\caption{Statistical parameters of data sets}
	\begin{tabular}{clll}
		\toprule
		Family	&Model	&Loss	&Sapce Complexity\\
		\midrule
    \multirow{6}*{\rotatebox{90}{Tensor models}}
		& DistMult	&$\bm{v}_h \times \bm{v}_r \times \bm{v}_t$, $\forall r\in \mathbb{R}$, $\bm{v}_r\ \text{is diagonal}$ 	&$\mathcal{O}(|\mathcal{E}|d+|\mathcal{R}|d)$	 \\

		& ComplEx &$\bm{v}_h\times \bm{v}_r\times \overline{\bm{v}_t}$, $\bm{v}_h$, $\bm{v}_t\in {{\mathbb{C}}^{d}}$; $\bm{v}_r\in {{\mathbb{C}}^{d\times d}}$	&$\mathcal{O}(|\mathcal{E}|d+|\mathcal{R}|d)$		 \\

    & Analogy &$\bm{v}_h \times \bm{v}_r\times \bm{v}_t$,
       $\forall r\in \mathbb{R}$,
       $\bm{v}_r\times \bm{v}_{r}^{\top }=\bm{v}_{r}^{\top }\times \bm{v}_r$;
       $\forall({r,\ 1},{r,\ 2})\in \mathbb{R}\times \mathbb{R}$,
       $\bm{v}_{r,\ 1}\times \bm{v}_{r,\ 2}=\bm{v}_{r,\ 2}\times \bm{v}_{r,\ 1}$ & $\mathcal{O}(|\mathcal{E}|d+|\mathcal{R}|d)$\\

    & SimplE & ${}^{1}/{}_{2}(\bm{v}_{h,\ h} \times r \times \bm{v}_{t,\ t})+{}^{1}/{}_{2}(\bm{v}_{h,\ t}\ \times \ \bm{v}_{r,\ -1} \times \bm{v}_{t,\ h})$,
    $\forall r\in \mathbb{R},\ \bm{v}_r,\ \bm{v}_{r,\ -1}\ \text{are diagonal} $ & $\mathcal{O}(2|\mathcal{E}|d+2|\mathcal{R}|d)$ \\

    & HolE & $(\bm{v}_h*\bm{v}_t)\times \bm{v}_r$ & $\mathcal{O}(|\mathcal{E}|d+|\mathcal{R}|d)$ \\

    & TruckER & $W\ {{\times }_{1}}\bm{v}_h\ {{\times }_{2}}\bm{v}_r\ {{\times }_{3}}\bm{v}_t$ & $\mathcal{O}(|\mathcal{E}|d_e+|\mathcal{R}|d_r+d_ed_rd_e)$ \\

    \midrule
    \multirow{4}*{\rotatebox{90}{Transform models}}

    & TransE & $\left\| \bm{v}_h+\bm{v}_r-\bm{v}_t \right\|$ & $\mathcal{O}(|\mathcal{E}|d+|\mathcal{R}|d)$ \\

    & CrossE & $\sigma (\tanh (\bm{v}_h\odot {{c}_{r}}+\bm{v}_r\odot \bm{v}_h\odot {{c}_{r}})\times {\bm{v}_{t}^{T}})$ & $\mathcal{O}(|\mathcal{E}|d+2|\mathcal{R}|d)$ \\

    & TorusE & $\text{mi}{{\text{n}}_{(x,y)\in ([\bm{v}_h]+[\bm{v}_r])\times [\bm{v}_t]}}{{\left\| \bm{v}_x-\bm{v}_y \right\|}_{i}}$ & $\mathcal{O}(|\mathcal{E}|d+|\mathcal{R}|d)$\\

    & RotatE & $-\left\| \bm{v}_h\odot \bm{v}_r-\bm{v}_t \right\|$,
    $\bm{v}_h,\bm{v}_r,\bm{v}_t\in {{\mathbb{C}}^{d}}$,
    $\forall \bm{v}_{r,\ i}\in \bm{v}_r$,
    $\left| \bm{v}_{r,\ i} \right|=1$ & $\mathcal{O}(|\mathcal{E}|d+|\mathcal{R}|d)$ \\

    \midrule
    \multirow{3}*{\rotatebox{90}{Deep models}}

    & ConvE &$g(W\times g([\bm{v}_h:\bm{v}_r]\otimes \omega )+b)\times \bm{v}_t$  &$\mathcal{O}(|\mathcal{E}|d+|\mathcal{R}|d+Tmn+Td(2{{d}_{m}}-m+1)({{d}_{n}}-n+1)) $ \\

    & ConvR &$g(W\times g([\bm{v}_h]\otimes {{\omega }_{r}})+b)\times \bm{v}_t$& $\mathcal{O}(|\mathcal{E}|{{d}_{e}}+|\mathcal{R}|{{d}_{r}}+Tmn+T{{d}_{e}}(2{{d}_{{{e}_{m}}}}-m+1)({{d}_{{{e}_{n}}}}-n+1))$ \\

    & RSN &$\sigma (\text{RSN}(\bm{v}_h,\bm{v}_r)\times \bm{v}_t)$ &$\mathcal{O}(2|\mathcal{E}|d+2|\mathcal{R}|d+Lknd)$ \\

    &HMAE&$g(Mul([\phi(\bar{\bm{v}_h});\bar{\bm{v}_r}]||[\bar{\bm{v}_h};\phi(\bar{\bm{v}_r})]))\bm{v}_t$ & $\mathcal{O}(|\mathcal{E}|d+|\mathcal{R}|d+Tmn+Td(2{{d}_{m}}-m+1)({{d}_{n}}-n+1)) $ \\

    &R-GCN& - & $\mathcal{O}(\mathcal{B}Kd^2+\mathcal{B}K|\mathcal{R}|)$ \\

    &CompGCN& -& $\mathcal{O}(Kd^2+\mathcal{B}d+\mathcal{B}|\mathcal{R}|)$ \\

		\bottomrule
	\end{tabular}
	\label{tab:statistic}
\end{table}
\end{landscape}

\noindent graph, and finally obtains the score of the fact through the linear layer and matrix multiplication with the entity embedding matrix.

\textbf{ConvR}\cite{ConvR} is also structured by a layer of convolutional neural network, in which the entities and relationships are represented as two vectors of different dimensions. Different from ConvE, ConvR decomposes the relationship vector into a set of convolution kernels.

\textbf{KMAE}\cite{KMAE} is a ConvE extension model that includes an entity kernel and a relation kernel in the input vectors, as well as a multi-attention mechanism in the convolutional process.

\textbf{RSN}\cite{RSN} is different from translational RNNs \blue{in} that the hidden state is \blue{updated by} re-using the head \blue{entity when} the input is a relationship. \blue{Almost like} ConvE, \blue{it computes the fact score by doing a dot product between the output vector and the target embedding.}

\textbf{R-GCN}\cite{R-GCN} is the fist \blue{GCN-based} knowledge representation model, which employs the GCN to deal with the influence of different edge relationships on nodes in the network structure. Different from the GCN\blue{, the R-GCN} considers the type and direction of edges.

\textbf{CompGCN}\cite{CompGCN}, which is built by two layers of multi-relational GCNs, leverages a \blue{variety} of entity-relationship composition operations from knowledge representation techniques to obtain the embeddings of entities and \blue{relaitonships}.

\subsection{Evaluation metrics}
According to Bordes et al\cite{TransE}, we calculate the scores of the corruption facts, which are generated by replacing \blue{an} entity \blue{in} the triplet \blue{in} the entity \blue{set}, and then rank the sequence index of the corresponding test triplet in descending order. Therefore, this paper also applies mean rank (MR), mean reciprocal \blue{rank (MRR)} and Hist@k, which are previously used \blue{in} the previous work as the evaluation metrics. It is worth noting that the experimental results are filtered by removing the scores of triplets \blue{in the} training set. Obviously, \blue{the lower the} MR and the \blue{higher the} Hist@k and MRR, the better the performance.

\subsection{Experimental setting}

\subsubsection{Hardware setting}
Pytorch is used to code our approach and train the model on workstations equipped with a Xeon 2.2GHz CPU, an NVIDIA 2080ti, and 128GB of memory.

\subsubsection{Hyperparameter setting}

The ranges of the \blue{hyperparameters} via the grid search are set as follows: learning rate {0.001, 0.002, 0.003, 0.005, 0.01}, batch size {64, 128, 256}, dropout rate {0.0, 0.1, 0.2, 0.3}. For dimensions, the initial embedding dimension is 200, and the graph convolutional \blue{embedding dimension is also} 200.
\blue{This paper employs the early stop strategy during the training process, in which each model is set to train 500 epochs, but training is terminated if the MRR on the valid set does not improve in 20 consecutive epochs.}

\subsection{Experimental results}

\begin{table}[h]
  \footnotesize
  \centering
	\caption{The link prediction results on FB15k-237 and WN18RR. Among them, best published scores in \underline{underlined} and best scores in \textbf{bold}. The results [$\star$] are taken from \cite{CompGCN}, the results [$\dagger$] are taken from \cite{KMAE}, and others are taken from \cite{Link_Prediction}.}
  \begin{tabular}{llcccccccc}
    \toprule
    \multirow{2}*{Family} &\multirow{2}*{\textbf{Model}} & \multicolumn{4}{c}{\textbf{FB15k-237}} & \multicolumn{4}{c}{\textbf{WN18RR}}\\
		\cmidrule(r){3-6}  \cmidrule(r){7-10}
		\multirow{2}*{} &\multirow{2}*{} & Hits@1 & Hits@10 & MR & MRR & Hits@1 & Hits@10 & MR & MRR \\
		\midrule
\multirow{6}*{{Tensor models}}
&DistMult [$\star$]	&15.5	&41.9	&254	&24.1	&39.	&49.	&5110	&43. \\
&ComplEx [$\star$]	&15.8	&42.8	&339	&24.7	&41.	&51.	&5261	&44. \\
&Analogy	&12.59	&35.38	&476	&20.2	&35.82	&38.00	&9266	&36.6 \\
&SimplE	&10.03	&34.35	&651	&17.9	&38.27	&42.65	&8764	&39.8 \\
&HolE	&21.37	&47.64	&186	&30.3	&40.28	&48.79	&8401	&43.2 \\
&TuckER	&25.90	&\underline{53.61}	&\underline{162}	&35.2	&42.95	&51.40	&6239	&45.9 \\
\midrule
\multirow{4}*{{Transform models}}
&TransE [$\dagger$]	&-	&46.5	&347	&29.4	&-	&50.1	&3384	&22.6 \\
&CrossE	&21.21	&47.05	&227	&29.8	&38.07	&44.99	&5215	&40.5 \\
&TorusE	&19.62	&44.71	&221	&28.1	&42.68	&53.35	&4873	&46.3 \\
&RotaE	&23.83	&53.06	&178	&33.6	&42.60	&\underline{57.35}	&\underline{3318}	&47.5 \\
\midrule
\multirow{6}*{{Deep models}}
&ConvE [$\star$]	&23.7	&50.1	&244	&32.5	&40.	&52.	&4187	&43. \\
&ConvR	&25.56	&52.63	&251	&34.6	&43.73	&52.68	&5646	&46.7 \\
&RSN	&19.84	&44.44	&248	&28.0	&34.59	&48.34	&210	&39.5 \\
&R-GCN [$\star$]	&15.8	&41.7	&-	&24.8	&-	&-	&-	&-\\
&CompGCN [$\star$]	&\underline{26.4}	&53.5	&197	&\underline{35.5}	&\underline{44.3}	&54.6	&3533	&\underline{47.9} \\
&KMAE [$\dagger$]	&24.	&50.2	&235	&32.6	&41.5	&52.4	&4441	&44.8\\
\midrule
&KBGSAT	&\textbf{34.45}	&\textbf{56.13}	&218	&\textbf{41.5}	&\textbf{48.2}	&57.1	&3485	&\textbf{50.9} \\
\bottomrule
\end{tabular}
	\label{tab3}
\end{table}

Table \ref{tab3} compares the performance of the GAT-based model KBGSAT, which has been retrained \blue{by a} semi-supervised self-training method, with the best published results of three \blue{types} of several state-of-the-art knowledge representation models from the literature on FB15k-237 and WN18RR benchmark datasets. The upper rows of the table are the results of tensor models, the middle rows show the results of translational distance models, followed by the results of deep neural network models, and the bottom row is the results of KBGSAT. Obviously, Table \ref{tab3} shows that the performances of KBGSAT are consistently ranked as the best on the metrics Hits@k and MRR over all competing basic methods on the benchmark dataset FB15k-237, and this approach has achieved significant improvements on some metrics of the \blue{dataset} WN18RR. \blue{On FB15k-237, KBGSAT achieves a 2.6\% higher Hits@10, a 6\% higher MRR, and a significant improvement of 34.5-26.4 = 8.1\% in Hits@1 (which is a 30.7\% relative improvement). Notably, when compared to previous SOTA on the benchmark data WN18RR, KBGSAT's performance has increased by at least 4\% on Hits@1 and 3\% on MRR. Notably, when compared to previous SOTA on the benchmark data WN18RR, KBGSAT's performance has increased by at least 4\% on Hits@1 and 3\% on MRR.}

\begin{figure}[h]
  \centering
  \includegraphics[width=0.95\textwidth]{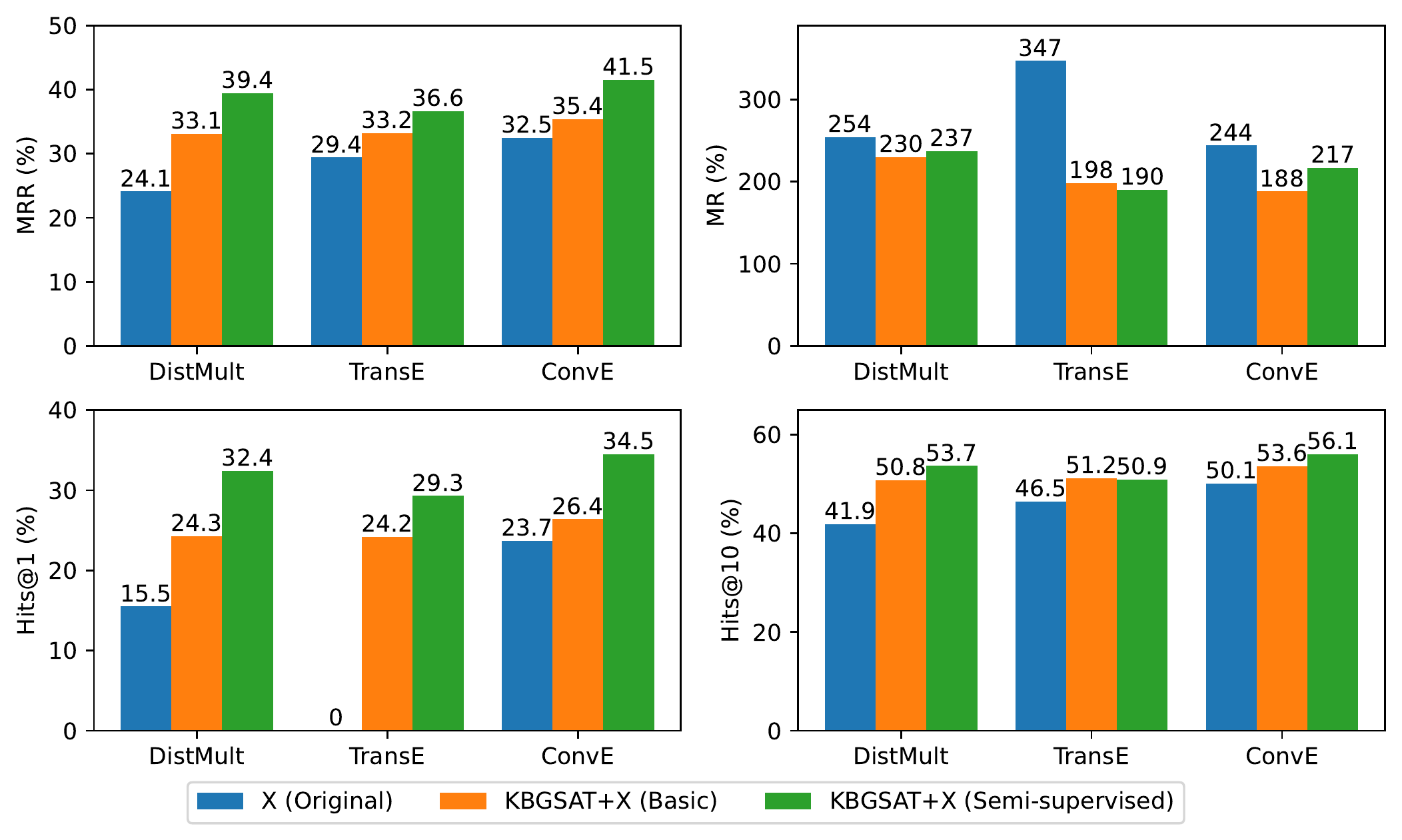}
  \caption{Compare the link prediction results of DistMult, TransE and ConvE on FB15k-237 under different conditions. Blue indicates the prediction results of the original models and the values from \cite{ConvE}(the value of TransE on Hits@1 is not given). \blue{The color orange} indicates that KBGSAT is used as the encoder, and these three models are the decoder. Green means that the semi-supervised self-training method has been used for retraining on the basis of orange. }
  \label{img3:fb237}

\end{figure}

For the four evaluation criteria, the larger the value of MRR and Hits@k, the better the performance of the model. Oppositely, the smaller the value of MR, the better the performance of the model. As shown in the figure \ref{img3:fb237}, on MRR, Hits@1 and Hits@10, the results of all three models show an upward trend. That means the encoder with KBGSAT structure and the semi-supervised self-training training method proposed in this paper can effectively improve the link prediction results on FB15k-237. On MR, the values have dropped significantly after using KBGSAT, but DistMult and ConvE have rallied after using the unsupervised training method. This is due to the presence of mislabeled negative samples in generating new positive samples during unsupervised training.

Table \ref{tab4} and \ref{tab5} show the link prediction results under basic and semi-unsupervised self-training \blue{conditions,} respectively. From \blue{the} view of the single table, most \blue{of the} results obtained by two-layer convolution are better on these two benchmark datasets. However, it is worth noting that the results \blue{obtained} by the KBGSAT with one-layer convolution \blue{are} significantly better than the other results of two-layer convolution\blue{ when} the decoder is DistMult on WN18RR. In addition, the numerical fluctuation of the results on the different \blue{scales} of convolution layers is very small, which \blue{indicates} that the multi-layer convolutions have little effectiveness \blue{in} improving the results. Therefore, this paper comprehensively considers \blue{using two-layer convolutions.}

\begin{landscape}
\begin{table}
  \footnotesize
  \centering
	\caption{Without the semi-supervised self-training process, the link prediction results of KBGSAT with different scale convolution layers on FB15k-237 and WN18RR.}
  \begin{tabular}{llcccccccccc}
    \toprule
    \multirow{2}*{N layers} &\multirow{2}*{\textbf{KBGSAT+X}} & \multicolumn{4}{c}{\textbf{FB15k-237}} & \multicolumn{4}{c}{\textbf{WN18RR}}\\
		\cmidrule(r){3-7}  \cmidrule(r){8-12}
		\multirow{2}*{} &\multirow{2}*{} &MRR &MR & Hits@1 &Hits@3 & Hits@10 & MRR & MR & Hits@1 & Hits@3 & Hits@10 \\
		\midrule
  \multirow{3}*{1}
  &DistMult	&0.3287	&214.1	&0.2404	&0.3604	&0.5060		&			0.43519	&3394.1	&0.39598	&0.444	&0.52074\\
	&TransE	&0.33481	&194.4	&0.24472	&0.37035	&0.51517		&			&&&&\\
	&ConvE	&0.3523	&195.3	&0.2622	&0.3856	&0.5319			&			0.47132	&3225.6	&0.43746	&0.48293	&0.53813\\

\midrule
\multirow{3}*{2}
  &DistMult	&0.33085	&229.6	&0.2427	&0.36282	&0.50787		&			0.41734	&5048.9	&0.38034	&0.42821	&0.49872\\
  &TransE	&0.33281	&198	&0.24167	&0.36795	&0.5118		&			&&&&\\
  &ConvE	&0.3555	&224.3	&0.2663	&0.3879	&0.5350		&			0.46717	&3380.1	&0.43267	&0.47735	&0.53829\\

    \bottomrule
    \end{tabular}
    	\label{tab4}
    \end{table}

    \begin{table}
      \footnotesize
      \centering
    	\caption{The link prediction results of KBGSAT with different scale convolution layers on FB15k-237 and WN18RR after the semi-supervised self-training process.}
      \begin{tabular}{llcccccccccc}
        \toprule
        \multirow{2}*{N layers} &\multirow{2}*{\textbf{KBGSAT+X}} & \multicolumn{4}{c}{\textbf{FB15k-237}} & \multicolumn{4}{c}{\textbf{WN18RR}}\\
    		\cmidrule(r){3-7}  \cmidrule(r){8-12}
    		\multirow{2}*{} &\multirow{2}*{} &MRR &MR & Hits@1 &Hits@3 & Hits@10 & MRR & MR & Hits@1 & Hits@3 & Hits@10 \\
    		\midrule
        \multirow{3}*{1}
        &DistMult	&0.38549	&208.2	&0.31364	&0.40413	&0.53354		&			0.46832	&3904.5	&0.4389	&0.47623	&0.53127\\
        &TransE	&0.36701	&190.5	&0.29224	&0.39253	&0.51312		&			&&&&\\
        &ConvE	&0.4101	&193.5	&0.34	&0.4282	&0.5536		&			0.51047	&3299.2	&0.48341	&0.51978	&0.55967\\

        \midrule
        \multirow{3}*{2}
        &DistMult	&0.39442	&237.1	&0.32432	&0.41378	&0.53694		&			0.44305	&5555.2	&0.41289	&0.45182	&0.50941\\
        &TransE	&0.36646	&190.3	&0.29312	&0.39084	&0.50906		&			&&&&\\
        &ConvE	&0.4154	&217.6	&0.3445	&0.4352	&0.5613		&			0.50863&	3484.8&	0.48227	&0.51595	&0.57063\\

        \bottomrule
        \end{tabular}
        	\label{tab5}
        \end{table}
\end{landscape}

\section{Conclusion}
For the edge structure characteristics of the triplet in the knowledge graph, this paper proposes a new graph self-attention convolutional model, i.e, KBGSAT, which has a more suitable self-attention mechanism \blue{to} the Transformer. Experimental results prove that \red{the proposed} mechanism effectively helps the encoder of the embedding model to capture the information from \blue{neighbor nodes, which} improves the representation ability of KBGSAT and its performance on the link prediction task.
In addition, inspired by the process of human associative memory, this paper also designs a semi-supervised self-training method to fine-tune the model after the traditional training process.
\blue{The core idea of this approach is that it employs the greedy strategy to generate additional unverified positive samples based on the pretrained models.
Experimental results \blue{show} that the semi-supervised self-training method can effectively improve the performance of knowledge representation models on the link prediction tasks.}

\blue{Although the KBGSAT and semi-supervised self-training method have achieved encouraging results for the link prediction, some works still remain for us to do: 1) strengthen the ability of graph convolutional layers to capture relationship features; 2) improve the ability of the semi-supervised self-training method on generating positive samples.} \newline

\noindent\textbf{\normalsize Acknowledgments}

This work is supported by National Science and Technology Innovation 2030-Key Project of "New Generation Artificial Intelligence" under Grant 2021ZD0113103, and the Natural Science Foundation of Jiangsu Provincial Higher Education under Grant 19KJB520008.

\bibliographystyle{elsarticle-num}
\bibliography{references}

\end{document}